\documentclass[runningheads]{llncs}
\pdfoutput=1
\let\llncssubparagraph\subparagraph
\let\subparagraph\paragraph
\usepackage[compact]{titlesec}
\let\subparagraph\llncssubparagraph

\usepackage{times}
\usepackage{amsmath}
\usepackage{amssymb}
\usepackage{graphicx}
\usepackage{named}
\usepackage{setspace}
\usepackage{listings}
\usepackage{url}
\usepackage{titlesec}

\long\def\BOC#1\EOC{\message{(Commented text )}}
\long\def\BOCC#1\EOCC{\message{(Commented text )}}
\long\def\BOCCC#1\EOCCC{\message{(Commented text )}}
\long\def\optional#1{\empty}

\long\def\NB#1{}

\def\sm{\hbox{\rm SM}}

\def\ar{\leftarrow}
\def\beq{\begin{equation}}
\def\eeq#1{\label{#1}\end{equation}}
\def\ba{\begin{array}}
\def\ea{\end{array}}
\def\i#1{\hbox{\it #1\/}}

\def\sm{\hbox{\rm SM}}

\def\no{\i{not}}

\def\ar{\ \leftarrow\ }
\def\rar{\rightarrow}
\def\lrar{\leftrightarrow}
\def\no{\i{not}}

\def\mvis{\!=\!}
\def\false{\hbox{\sc false}}
\def\true{\hbox{\sc true}}

\def\i#1{\hbox{\itshape #1\/}}

\DeclareSymbolFont{AMSa}{U}{msa}{m}{n}
\DeclareMathSymbol{\square}{\mathord}{AMSa}{"03}

\def\wh{\widehat}
\def\mvis{\!=\!}
\def\sneg{\sim\!\!}

\def\bX{{\bf{x}}}
\def\bC{{\bf{c}}}
\def\vbC{{\wh{\bf{c}}}}

\def\cbl{\hbox{\rm CBL}}

\def\bfxi{\boldsymbol{\xi}}

\newtheorem{prop}{Proposition}
\newtheorem{thm}{Theorem}
\newtheorem{cor}{Corollary}

\newtheorem{example1}{Example}

\begin{document}

\mainmatter

\title{A Functional View of Strong Negation \\in Answer Set Programming}

\author{Michael Bartholomew and Joohyung Lee}

\institute{
School of Computing, Informatics, and Decision Systems Engineering \\
Arizona State University, Tempe, USA
}

\titlerunning{A Functional View of Strong Negation in Answer Set Programming}
\authorrunning{M.~Bartholomew and J.~Lee}
\setcounter{page}{49}

\maketitle

\begin{abstract}
The distinction between strong negation and default negation has been
useful in answer set programming. We present an alternative account of
strong negation, which lets us view strong negation in terms of the
functional stable model semantics by Bartholomew and Lee. More
specifically, we show that, under complete interpretations, minimizing
both positive and negative literals in the traditional answer set
semantics is essentially the same as ensuring the uniqueness of
Boolean function values under the functional stable model semantics. 
The same account lets us view Lifschitz's two-valued logic programs as
a special case of the functional stable model semantics. 
In addition, we show how non-Boolean intensional functions can be
eliminated in favor of Boolean intensional functions, and
furthermore can be represented using strong negation, 
which provides a way to compute the functional stable model
semantics using existing ASP solvers. We also note that similar
results hold with the functional stable model semantics by Cabalar.
\end{abstract}

\section{Introduction}

The distinction between default negation and strong negation has been
useful in answer set programming. In particular, it
yields an elegant solution to the frame problem. The fact that block
$b$ stays at the same location $l$ by inertia can be described by the
rule  
\beq
\ba l
  \i{On}(b,l,t\!+\!1)\ar \i{On}(b,l,t),\ \no\ \sneg\i{On}(b,l,t\!+\!1)
\ea
\eeq{inertia-on-sneg}
along with the rule that describes the uniqueness of location values
\cite{lif02},
\beq
  \sneg\i{On}(b,l_1,t)\ar\i{On}(b,l,t),\ l\ne l_1\ .
\eeq{unique-on-sneg}
Here `$\sneg\ $' is the symbol for strong negation that represents
explicit falsity while `$\no$' is the symbol for default negation
(negation as failure). 
Rule~(\ref{inertia-on-sneg}) asserts that without explicit evidence to the
contrary, block $b$ remains at location $l$.  If we are given explicit
conflicting information about the location of $b$ at time $t\!+\!1$
then this conclusion will be defeated by rule~(\ref{unique-on-sneg}),
which asserts the uniqueness of location values.

An alternative representation of inertia, which uses choice rules
instead of strong negation, was recently presented by
\citeauthor{bartholomew12stable}~[\citeyear{bartholomew12stable}].
Instead of rule~\eqref{inertia-on-sneg}, they use the choice rule
\beq
 \{\i{On}(b,l,t\!+\!1)\} \ar\i{On}(b,l,t)\ , 
\eeq{inertia-on-choice}
which states that ``if $b$ is at $l$ at time $t$, then decide
arbitrarily whether to assert that $b$ is at $l$ at time  $t\!+\!1$.'' 
Instead of rule~\eqref{unique-on-sneg}, they write weaker rules 
for describing the functional property of $\i{On}$: 
\begin{align}
   \ar \{\i{On}(b,l,t): Location(l)\} 0 &  
        \qquad \text{(existence of location)} \label{exists-on}  \\
   \ar 2\{\i{On}(b,l,t): Location(l)\}\   & 
        \qquad \text{(uniqueness of location)}, \label{unique-on}  
\end{align}
which can be also combined into one rule:
$
   \ar \no\ 1\{\i{On}(b,l,t): Location(l)\} 1\ .
$
In the absence of additional information about the location of block
$b$ at time $t\!+\!1$, asserting $\i{On}(b,l,t\!+\!1)$ is the only
option, in view of the existence of location constraint
\eqref{exists-on}.  But if we
are given conflicting information about the location of $b$ at time
$t\!+\!1$ then not asserting $\i{On}(b,l,t\!+\!1)$ is the only option,
in view of the uniqueness of location constraint \eqref{unique-on}. 

Rules~\eqref{inertia-on-choice}, \eqref{exists-on}, and
\eqref{unique-on} together can be more succinctly represented in
the language of~\cite{bartholomew12stable} by means of intensional
functions. That is, the three rules can be replaced by 
one rule
\beq
  \{\i{Loc}(b,t\!+\!1)=l\}\ar \i{Loc}(b,t)=l\ ,
\eeq{inertia-loc-choice}
where $\i{Loc}$ is an intensional function constant (the rule reads,
``if block $b$ is at location $l$ at time $t$, by default, the block
is at $l$ at time $t\!+\!1$'').
In fact, Corollary~2 of \cite{bartholomew12stable} tells us how to
eliminate intensional functions in favor of intensional predicates,
justifying the equivalence between~\eqref{inertia-loc-choice} and the 
set of rules~\eqref{inertia-on-choice}, \eqref{exists-on}, and
\eqref{unique-on}. 
The translation allows us to compute the language
of~\cite{bartholomew12stable} using existing ASP solvers, such as {\sc
  smodels} and {\sc gringo}. However, {\sc dlv} cannot be used because
it does not accept choice rules. On the other hand, all these solvers
accept rules~\eqref{inertia-on-sneg} and \eqref{unique-on-sneg}, which
contain strong negation.

The two representations of inertia involving intensional predicate
$\i{On}$ do not result in the same answer sets. In the first
representation, which uses strong negation, each answer set contains
only one atom of the form $\i{On}(b,l,t)$ for each block $b$ and each
time $t$; for all other locations $l'$, negative literals
$\sneg\i{On}(b,l',t)$ belong to the answer set. 
On the other hand, such negative literals do not occur in the answer
sets of a program that follows the second representation, which
yields fewer ground atoms. 
This difference can be well explained
by the difference between the symmetric and the asymmetric views of
predicates that
Lifschitz described in his message to Texas Action Group, titled
``Choice Rules and the Belief-Based View of
ASP'':~\footnote{http://www.cs.utexas.edu/users/vl/tag/choice\_discussion}
\begin{quote} 
The way I see it, in ASP programs we use predicates of two kinds,
let's call them ``symmetric'' and ``asymmetric.''  The fact that an
object $a$ does not have a property~$p$ is reflected by the presence of
$\sneg p(a)$ in the answer set if $p$ is ``symmetric,'' and by the
absence of $p(a)$ if $p$ is ``asymmetric.''  In the second case, the
strong negation of $p$ is not used in the program at all. 
\end{quote} 

According to these terminologies, predicate $\i{On}$ is symmetric in
the first representation, and asymmetric in the second
representation. 

This paper presents several technical results that help us understand
the relationship between these two views. In this regard, it helps us
to understand strong negation as a way of expressing intensional Boolean
functions. 

\begin{itemize}
\item  Our first result provides an alternative account of strong
  negation in terms of Boolean intensional functions. For instance, 
  \eqref{inertia-on-sneg} can be identified with 
  \[ 
    \i{On}(b,l,t\!+\!1)\mvis\true\ar \i{On}(b,l,t)\mvis\true \land \neg
    (\i{On}(b,l,t\!+\!1)\mvis\false)\ ,
  \] 
  and \eqref{unique-on-sneg} can be identified with
  \[ 
    \i{On}(b,l_1,t)\mvis\false\ar\i{On}(b,l,t)\mvis\true\land l\ne l_1\ .
  \] 
  Under complete interpretations, we show that minimizing both
  positive and negative literals in the traditional answer set
  semantics is essentially the same as ensuring the uniqueness of
  Boolean function values under the functional stable model semantics. 
  In this sense, strong negation can be viewed as a mere
  disguise of Boolean functions.\footnote{It is also well-known that
    strong negation can be also viewed in terms of auxiliary predicate
  constants \cite{gel91b}.}
  
\item  We show how non-Boolean intensional functions can be eliminated
  in favor of Boolean functions.
  Combined with the result in the first bullet, this tells us a new
  way of turning the language of \cite{bartholomew12stable} into
  traditional answer set programs with strong negation, so that system
  {\sc dlv}, as well as {\sc smodels} and {\sc gringo}, can be used
  for computing the language of \cite{bartholomew12stable}. 
  As an example, it tells us how to turn~\eqref{inertia-loc-choice}
  into the set of rules~\eqref{inertia-on-sneg} and
  \eqref{unique-on-sneg}.

\item  \citeauthor{lifschitz12two-valued}
  [\citeyear{lifschitz12two-valued}] recently proposed  ``two-valued
  logic programs,'' which modifies the traditional stable model
  semantics to represent complete information without
  distinguishing between strong negation and default negation. Using
  our result that views strong negation in terms of Boolean functions,
  we show that two-valued logic programs are in fact a special case of
  the functional stable model semantics in which every function is
  Boolean. 

\end{itemize}

While the main results are stated for the language
of~\cite{bartholomew12stable}, similar results hold with the language
of~\cite{cabalar11functional} based on the relationship between the
two languages studied in~\cite{bartholomew13onthestable}. Furthermore,
we note that the complete interpretation assumption in the first
bullet can be dropped if we instead refer to the language
of~\cite{cabalar11functional}, at the price of introducing partial
interpretations. 

The paper is organized as follows. In Section~\ref{sec:preliminaries}
we review the two versions of the stable model semantics, one that
allows strong negation, but is limited to express intensional
predicates only, and the other that allows both intensional predicates
and intensional functions. As a special case of the latter
we also present multi-valued propositional formulas under the stable
model semantics. 
Section~\ref{sec:sneg-bf} shows how strong negation can be viewed in
terms of Boolean functions. 
Section~\ref{sec:nonbool-sneg} shows how non-Boolean functions can be
eliminated in favor of Boolean functions. Section~\ref{sec:tvlp} shows
how Lifschitz's two-valued logic programs can be viewed as a special
case of the functional stable model
semantics. Section~\ref{sec:sneg-cabalar} shows how strong negation
can be represented in the language of 
\cite{cabalar11functional}.


\section{Preliminaries}\label{sec:preliminaries}



\subsection{Review: First-Order Stable Model Semantics and Strong
  Negation}
  \label{ssec:review-sneg}

This review follows~\cite{ferraris11stable}. 
A {\em signature} is defined as in first-order logic, consisting of 
 {\em function constants} and {\em predicate
  constants}. Function constants of arity $0$ are also called {\em
  object constants}.
We assume the following set of primitive propositional connectives
and quantifiers:
$\bot\hbox{ (falsity)},\ \land,\ \lor,\ \rar,\ \forall,\ \exists$.
The syntax of a formula is defined as in first-order logic. We
understand $\neg F$ as an abbreviation of $F\rar\bot$.

The stable models of a sentence $F$ relative to a list of predicates
${\bf p} = (p_1,\dots,p_n)$ are defined via the {\em stable model
  operator with the intensional predicates ${\bf p}$}, denoted by
$\sm[F; {\bf p}]$.
Let ${\bf u}$ be a list of distinct predicate variables
$u_1,\dots,u_n$ of the same length as~${\bf p}$.
By ${\bf u}={\bf p}$ we denote the conjunction of the formulas
$\forall {\bf x}(u_i({\bf x})\lrar p_i({\bf x}))$, where ${\bf x}$ is a
list of distinct object variables of the same length as the arity of
$p_i$, for all $i=1,\dots, n$.  By ${\bf u}\leq{\bf p}$ we denote the
conjunction of the formulas
$\forall {\bf x}(u_i({\bf x})\rar p_i({\bf x}))$ for all
$i=1,\dots, n$, and
${\bf u}<{\bf p}$ stands for
$({\bf u}\leq{\bf p})\land\neg({\bf u}={\bf p})$.
For any first-order sentence $F$, expression $\sm[F;{\bf p}]$ stands
for the second-order sentence
\[ 
   F \land \neg \exists {\bf u} (({\bf u}<{\bf p}) \land F^*({\bf u})),
\] 
where
$F^*({\bf u})$ is defined recursively:
\begin{itemize}
\item  $p_i({\bf t})^* = u_i({\bf t})$ for any list ${\bf t}$ of terms;
\item  $F^* = F$ for any atomic formula~$F$ (including $\bot$ and
  equality)        that does not contain members of~${\bf p}$;
\item  $(F\land G)^* = F^* \land G^*$;\ \ \ \ \  $(F\lor G)^* = F^* \lor G^*$;
\item  $(F\rar G)^* = (F^* \rar G^*)\land (F \rar G)$;
\item  $(\forall xF)^* = \forall xF^*$;\ \ \ \ \ $(\exists xF)^* =
  \exists x F^*$.
\end{itemize}

A model of a sentence $F$ (in the sense of first-order logic) is
called {\em ${\bf p}$-stable} if it satisfies $\sm[F; {\bf p}]$.

The traditional stable models of a logic program $\Pi$ are identical
to the Herbrand stable models of the {\em FOL-representation} of $\Pi$
(i.e., the conjunction of the universal closures of implications
corresponding to the rules).

\BOCC
\begin{example1}
For program $\Pi$ that contains three rules
\[
\ba l
  p(a) \qquad\qquad  q(b)\qquad\qquad
  r(x)\ar p(x), \no\ q(x)
\ea
\]
the FOL-representation $F$ of $\Pi$ is
\beq
  p(a)\land q(b)\land\forall x((p(x)\land\neg q(x))\rar r(x))
\eeq{ex3f}
and $\sm[F]$ is
$$
\ba l
p(a)\land q(b)\land\forall x((p(x)\land\neg q(x))\rar r(x))\\
\quad\land\neg\exists uvw(((u,v,w)<(p,q,r))\wedge u(a)\land v(b) \\
\hspace{5em} \land\forall x(((u(x)\land
(\neg v(x)\land\neg q(x))
)\rar w(x))\land((p(x)\land\neg q(x))\rar r(x)))),
\ea
$$
which is equivalent to first-order sentence
\beq
\ba l
\forall x(p(x) \lrar x=a) \land \forall x(q(x) \lrar x=b)
\land \forall x (r(x) \lrar (p(x) \land \neg q(x)))
\ea
\eeq{ex3f-comp} (see \cite{fer07a}, Example~3). The stable models of
$F$ are any first-order models of (\ref{ex3f-comp}). The only Herbrand
stable models of $F$ is  $\{p(a),\ q(b),\ r(a)\}$.
\end{example1}
\EOCC

\citeauthor{ferraris11stable} [\citeyear{ferraris11stable}]
incorporate strong negation into the stable model semantics by
distinguishing between intensional predicates of two kinds, {\sl
  positive} and {\sl negative}. Each negative intensional predicate
has the form  $\sneg p$, where $p$ is a positive intensional predicate
and `$\sneg$\ ' is a  symbol for strong negation. In this sense,
syntactically
$\sneg\ $ is not a logical connective, as it can appear only as a part
of a predicate constant. An interpretation of the underlying signature
is {\sl coherent} if it satisfies the formula
$ \neg\exists {\bf x}(p({\bf x})\,\land\sneg p({\bf x}))$,
where~${\bf x}$ is a list of distinct object variables, for each
negative predicate~$\sneg p$. We consider coherent interpretations
only.

\begin{example1}\label{ex:bw-sneg0}
The following is a representation of the Blocks World in the syntax of
logic programs:
\beq
\small
\ba {rcll}
  \bot&\ar&\i{On}(b_1,b,t), \i{On}(b_2,b,t) &  (b_1\neq b_2) \\
  \i{On}(b,l,t+1) &\ar&\i{Move}(b,l,t) \\
  \bot&\ar&\i{Move}(b,l,t), \i{On}(b_1,b,t) \\
  \bot&\ar&\i{Move}(b,b_1,t), \i{Move}(b_1,l,t) \\
  \i{On}(b,l,0) &\ar& \no\ \sneg \i{On}(b,l,0) \\
  \sneg\i{On}(b,l,0) &\ar& \no\ \i{On}(b,l,0) \\
  \i{Move}(b,l,t) &\ar& \no\ \sneg \i{Move}(b,l,t) \\
  \sneg\i{Move}(b,l,t) &\ar& \no\ \i{Move}(b,l,t) \\
  \i{On}(b,l,t+1)&\ar&\i{On}(b,l,t), \no\ \sneg\i{On}(b,l,t+1) \\
  \sneg\i{On}(b,l,t)&\ar& \i{On}(b,l_1,t)  & (l\ne l_1)\ .
\ea 
\eeq{bw-sneg0}
Here $\i{On}$ and $\i{Move}$ are intensional predicate constants, $b$, $b_1$,
$b_2$ are variables ranging over the blocks, $l$, $l_1$ are variables
ranging over the locations (blocks and the table), and $t$ is a
variable ranging over the timepoints. The first rule asserts that at
most one block can be on another block. The next three rules describe the effect
and preconditions of action $\i{Move}$. The next four rules describe
that fluent $\i{On}$ is initially exogenous, and action
$\i{Move}$ is exogenous at each time. The next rule describes 
inertia, and the last rule asserts that a block can be at most at one 
location.
\end{example1}

\NB{check the program} 


\subsection{Review:  The Functional Stable Model Semantics}
  \label{ssec:review-fsm}

The functional stable model semantics is defined by modifying the
semantics in the previous section to allow ``intensional'' functions 
\cite{bartholomew12stable}. 
For predicate symbols (constants or variables) $u$ and $c$, 
we define $u\le c$ as 
\hbox{$\forall {\bf x}(u({\bf x})\rar c({\bf x}))$}.
We define $u=c$ as 
$\forall {\bf x}(u({\bf x})\lrar c({\bf x}))$
if $u$ and $c$ are predicate symbols, and 
$\forall {\bf x}(u({\bf x})=c({\bf x}))$ 
if they are function symbols. 

Let $\bC$ be a list of distinct predicate and function constants and
let  $\vbC$ be a list of distinct predicate and function variables
corresponding to~$\bC$. We call members of $\bC$ {\sl intensional}
constants.
By $\bC^{pred}$ we mean the list of the predicate constants in $\bC$,
and by $\vbC^{pred}$ the list of the corresponding predicate variables
in $\vbC$.
We define $\vbC<\bC$ as 
$ 
  (\vbC^{pred}\le \bC^{pred})\land\neg (\vbC=\bC)
$ 
and $\sm[F;\ \bC]$ as 
\[
   F\land\neg\exists \vbC(\vbC<\bC\land F^*(\vbC)),
\]
where $F^*(\vbC)$ is defined the same as the one in
Section~\ref{ssec:review-sneg} except for the base case: 
\begin{itemize}
\item When $F$ is an atomic formula, $F^*$ is $F'\land F$, where $F'$
  is obtained from $F$ by replacing all intensional (function and
  predicate) constants in it with the corresponding (function and
  predicate) variables.
\end{itemize}


If ${\bf c}$ contains predicate constants only, this definition of a
stable model reduces to the one in~\cite{ferraris11stable}, also
reviewed in Section~\ref{ssec:review-sneg}. 

According to~\cite{bartholomew12stable}, a {\em choice formula}
$\{F\}$ is an abbreviation of the formula
$F\lor\neg F$, which is also strongly equivalent to $\neg\neg F\rar
F$. 
A formula $\{{\bf t} = {\bf t}'\}$, where ${\bf t}$ contains an
intensional function constant and ${\bf t}'$ does not, represents that
${\bf t}$ takes the value ${\bf t}'$ by default, as the following
example demonstrates.

\begin{example1}\label{ex:f=1}
Let $F_1$ be $\{f=1\}$, which stands for 
$(f\mvis 1)\lor\neg (f\mvis 1)$, and $I_1$ be an interpretation such
that $I_1(f)=1$. Let's assume that we consider only
interpretations that map numbers to themselves. $I_1$ is an $f$-stable
model of $F_1$:  $F_1^*(\wh{f})$ is equivalent to 
$((\wh{f}\mvis 1)\land (f\mvis 1))\lor\neg (f\mvis 1)$,\footnote{%
{\rm It holds that $(\neg F)^*$ is equivalent to $\neg F$}.} 
which is further equivalent to $(\wh{f}\mvis 1)$ under the assumption
$I_1$. It is not possible to satisfy this formula by assigning
$\wh{f}$ a different value from $I_1(f)$. On the other hand, $I_2$
such that $I_2(f)=2$ is not $f$-stable since $F_1^*(\wh{f})$ is
equivalent to $\top$ under $I_2$, so that it is possible to satisfy
this formula by assigning $\wh{f}$ a different value from $I_2(f)$.
If we let $F_2$ be $\{f=1\}\land (f=2)$, then $I_2$ is a $f$-stable of
$F_2$, but $I_1$ is not:
$F_2^*(\wh{f})$ is equivalent to $\wh{f}\mvis 2$ under $I_2$, so that
$\wh{f}$ has to map to $2$ as well. This example illustrates the
nonmonotonicity of the semantics. 
\end{example1}

\begin{example1}\label{ex:bw-func}
The Blocks World can be described in this language as follows. For
readability, we write in a logic program like syntax: 
\[
\small
\ba {rcl}
  \bot&\ar&\i{Loc}(b_1,t) \mvis b\land\i{Loc}(b_2,t) \mvis b \land 
  (b_1\ne b_2) \\
  \i{Loc}(b,t\!+\!1) \mvis l &\ar& \i{Move}(b,l,t) \\
  \bot &\ar& \i{Move}(b,l,t) \land \i{Loc}(b_1,t) \mvis b \\
  \bot &\ar& \i{Move}(b,b_1,t) \land \i{Move}(b_1,l,t)  \\
  \{\i{Loc}(b,0) \mvis l\} \\
  \{ \i{Move}(b,l,t)\} \\
  \{\i{Loc}(b,t\!+\!1) \mvis l\}&\ar& \i{Loc}(b,t) \mvis l \ .
\ea
\]
Here $\i{Loc}$ is a function constant. The last rule is a default
formula that describes the commonsense law of inertia. The stable
models of this program are the models of $\sm[F;\ \i{Loc}, \i{Move}]$,
where $F$ is the FOL-representation of the program. 
\end{example1}


\NB{Theorem X from XX states that stable models of multi-valued
  propositional formulas is a special case.}


\subsection{Review: Stable Models of Multi-Valued Propositional Formulas}\label{ssec:sm-mvpf}

The following is a review of the stable model semantics of
multi-valued propositional formulas from~\cite{bartholomew12stable}, 
which can be viewed as a special case of the functional stable model
semantics in the previous section.

The syntax of multi-valued propositional formulas is given
in~\cite{ferraris11stable}. 
A {\sl multi-valued propositional signature} is a set $\sigma$~of
symbols called {\sl constants}, along with a nonempty finite
set~$\i{Dom}(c)$ of symbols, disjoint from $\sigma$, assigned to each
constant~$c$.  We call $\i{Dom}(c)$ the {\sl domain} of~$c$.
A {\sl Boolean} constant is one whose domain is the
set~${\{\true, \false\}}$. 
An {\sl atom} of a signature~$\sigma$ is an expression of the form
${c\mvis v}$ (``the value of~$c$ is~$v$'') where $c \in \sigma$ and $v
\in \i{Dom}(c)$. 
A {\sl (multi-valued propositional) formula} of~$\sigma$ is a
propositional combination of atoms.

A {\sl (multi-valued propositional) interpretation} of~$\sigma$ is a function that maps every
element of~$\sigma$ to an element of its domain.  An interpretation~$I$
{\sl satisfies} an atom ${c\mvis v}$ (symbolically, ${I\models c\mvis v}$)
if $I(c)=v$.
The satisfaction relation is extended from atoms to arbitrary
formulas according to the usual truth tables for the propositional
connectives. $I$ is a {\em model} of a formula if it satisfies the
formula.

The reduct $F^I$ of a multi-valued propositional formula $F$ relative
to a multi-valued propositional interpretation $I$ is the formula
obtained from $F$ by replacing each maximal subformula that is not
satisfied by $I$ with $\bot$. Interpretation $I$ is a {\sl stable
  model} of $F$ if $I$ is the only interpretation satisfying $F^I$.

\BOCC
\begin{example1}\label{ex6}
Consider a multi-valued propositional signature
$\sigma = \{\i{ClrBlue},\i{\i{ClrRed}},\i{TapeClr}\}$, where
$\i{Dom}(\i{ClrBlue})=\i{Dom}(\i{ClrRed})=\{\true,\false\}$ and 
$\i{Dom}(\i{TapeClr})=\{\i{Red},\i{Blue},\i{Green}\}$. The following
is a multi-valued propositional formula $F$: \\[-0.5em]
\[
\ba l
    (\i{ClrBlue}=\true \lor \i{ClrBlue}=\false) 
   \land\ (\i{ClrRed}=\true \lor \i{ClrRed}=\false) \\
  \land\ (\i{ClrBlue}=\true \rightarrow \i{TapeClr}=\i{Blue})
   \land\ (\i{ClrRed}=\true \rightarrow \i{TapeClr}=\i{Red})\ . 
\ea
\]
Consider an interpretation $I$ such that 
$I(\i{ClrBlue})=\false$, $I(\i{ClrRed})=\true$
and $I(\i{TapeClr})=\i{Red}$. 
The reduct $F^I$ is \\[-0.5em]
\[
\ba l
  (\bot\lor \i{ClrBlue}\mvis \false) 
  \land (\i{ClrRed}\mvis \true\lor \bot) \\
 \land\ (\bot \rightarrow \bot)
  \land (\i{ClrRed}\mvis \true \rightarrow \i{TapeClr}\mvis \i{Red}),  
\ea
\]
and $I$ is the only interpretation of $\sigma$ that satisfies $F^I$.
\end{example1}
\EOCC

\begin{example1}
Similar to Example~\ref{ex:f=1}, consider the signature $\sigma=\{f\}$
such that
$\i{Dom}(c)=\{1,2,3\}$. Let $I_1$ be an interpretation such that
$I_1(c)=1$, and $I_2$ be such that $I_2(c)=2$. 
Recall that $\{f\mvis 1\}$ is shorthand for $(f\mvis 1)\lor \neg
(f\mvis 1)$. The reduct of this formula relative to $I_1$ is
$(f\mvis 1)\lor\bot$, and $I_1$ is the only model of the
reduct. 
On the other hand, the reduct of $\{f\mvis 1\}$
relative to $I_2$ is $(\bot\lor\neg\bot)$ and $I_2$ is not its unique
model. 
Also, the reduct of $\{f\mvis 1\}\land (f\mvis 2)$ relative to $I_1$
is $(\bot\lor\neg\bot)\land \bot$ and $I_1$ is not a model. The reduct
of $\{f\mvis 1\}\land (f\mvis 2)$ relative to $I_2$ is
$(\bot\lor\neg\bot)\land (f\mvis 2)$, and $I_2$ is the only
model of the reduct. 
\end{example1}
\bigskip


\BOCC
\subsection{Multi-Valued Propositional Formulas and FO Formulas}

Multi-valued propositional formulas can be identified with a special
case of first-order formulas as follows. 
Let $F$ be a multi-valued propositional formula of signature~$\sigma$.
We identify $\sigma$ with a first-order signature $\sigma'$ that consists of
\begin{itemize}
\item  all symbols from $\sigma$ as object constants, and
\item  all symbols from $\i{Dom}(c)$ where $c$ is in $\sigma$ as
  object constants. 
\end{itemize}

We may view multi-valued propositional interpretations of $\sigma$
as a special case of first-order interpretations of~$\sigma'$. We
say that a first-order interpretation $I$ of $\sigma'$ {\sl conforms}
to $\sigma$ if
\begin{itemize}
\item  the universe of $I$ is the union of $\i{Dom}(c)$ for all $c$ in
  $\sigma$;  
\item  $c^I\in\i{Dom}(c)$ for every $c$ in $\sigma$; 
\item  $v^I = v$ for every $v$ in $\i{Dom}(c)$ where $c\in\sigma$.

\end{itemize}

\begin{prop}\label{prop:mvpf-fo} \optional{prop:mvpf-fo}
For any multi-valued propositional formula $F$ of signature
$\sigma$ such that \hbox{$|\i{Dom}(c)|\ge 2$} for every $c\in\sigma$,
an interpretation $I$ of $\sigma$ is a multi-valued
propositional stable model of $F$ iff $I$ is an interpretation of
$\sigma'$ that conforms to $\sigma$ and satisfies $\sm[F; \sigma]$. 
\end{prop}
\EOCC


\vspace{-0.5cm}
\section{Relating Strong Negation to Boolean
  Functions}\label{sec:sneg-bf}


\vspace{-0.2cm}
\subsection{Representing Strong Negation in Multi-Valued Propositional
Formulas}


Given a traditional propositional logic program $\Pi$ of a signature
$\sigma$ \cite{gel91b}, we identify $\sigma$ with the multi-valued propositional
signature whose constants are the same symbols from $\sigma$ and every
constant is Boolean. 
By $\Pi^{mv}$ we mean the
multi-valued propositional formula that is obtained from $\Pi$ by
replacing negative literals of the form $\sneg p$ with $p=\false$ and
positive literals of the form $p$ with $p=\true$.

We say that a set $X$ of literals from~$\sigma$ is {\sl complete} if, 
for each atom $a\in\sigma$, either $a$ or $\sneg a$ is in $X$.
We identify a complete set of literals from $\sigma$ with the
corresponding multi-valued propositional interpretation. 

\begin{thm}\label{thm:sneg-bool-prop}\optional{thm:sneg-bool-prop}
A complete set of literals is an answer set of $\Pi$ in the sense
of~\cite{gel91b} iff it is a stable model of $\Pi^{mv}$ in the sense
of~\cite{bartholomew12stable}. 
\end{thm}

The theorem tells us that checking the minimality of positive and
negative literals under the traditional stable model semantics
is essentially the same as checking the uniqueness of corresponding
function values under the stable model semantics
from~\cite{bartholomew12stable}. 

\begin{example1}\label{ex:simple}
Consider the program that describes a simple transition system
consisting of two states depending on whether fluent $p$ is true or
false, and an action that makes $p$ true (subscripts $0$ and $1$
represent time stamps).  \\[-0.5em]
\beq
\small
\ba {rrclrcl}
\hspace{3em} & p_0 &\ar& \no\ \sneg p_0  \hspace{5em}&   p_1 &\ar& a \\ 
 & \sneg p_0 &\ar& \no\ p_0 \\  
 &           &   &          & p_1&\ar& p_0,\no\ \sneg p_1 \\
 & a &\ar& \no\ \sneg a &  \sneg p_1&\ar& \sneg p_0, \no\ p_1 \ . \\
 & \sneg a &\ar& \no\ a 
\ea 
\eeq{simple-sneg}
The program has four answer sets, each of which corresponds to one of
the four edges of the transition system. For instance, $\{\sneg p_0,
a, p_1\}$ is an answer set.  This program can be encoded in the input
languages of {\sc gringo}  and {\sc dlv}. In the input language of {\sc
  dlv}, which allows disjunctions in the head of a rule,  the four rules in the
first column can be succinctly replaced by  \\[-0.5em]
\[
\small
   p_0\lor \sneg p_0 \hspace{5em}      a \lor \sneg a\ .
\]

According to Theorem~\ref{thm:sneg-bool-prop}, the stable
models of this program are the same as the stable models of the
following multi-valued propositional formula (written in a logic
program style syntax; `$\neg$' represents default negation): \\[-3.8em]

\begin{multicols}{2}
\[
\small
\ba {rcl} 
  p_0\mvis\true &\ar& \neg (p_0\mvis\false)  \\
  p_0\mvis\false &\ar& \neg (p_0\mvis\true) \\ \\

  a\mvis\true &\ar& \neg (a\mvis\false) \\
  a\mvis\false &\ar& \neg (a\mvis\true)
\ea
\]

\[
\small
\ba {rcl}
  p_1\mvis\true &\ar& a\mvis\true \\ \\

  p_1\mvis\true &\ar& p_0\mvis\true \land \neg (p_1=\false) \\
  p_1\mvis\false &\ar& p_0\mvis\false \land \neg (p_1=\true)\ .  \\ \\
\ea 
\] 
\end{multicols}
\end{example1}

\NB{Show simple-sneg, then simple-mv, then simple-fn-pred (compare
  with simple-sneg (later with blocks world?}


\subsection{Relation among Strong Negation, Default Negation, Choice
  Rules and Boolean Functions}

In certain cases, strong negation can be replaced by default negation,
and furthermore the expression can be rewritten in terms of choice
rules, which often yields a succinct representation.

The following theorem, which extends the {\em Theorem on Double
  Negation} from~\cite{ferraris09symmetric} to allow intensional
functions, presents a condition under
which equivalent transformations in classical logic preserve stable
models. 

\begin{thm}\label{thm:dneg}\optional{thm:dneg}
Let $F$ be a sentence, let ${\bf c}$ be a list of predicate and
function constants, and let $I$ be a (coherent) interpretation.
Let  $F'$ be the sentence obtained from $F$ by replacing a subformula
$\neg H$ with $\neg H'$ such that  
$I\models \widetilde {\forall} (H\lrar H')$. 
Then
\[ 
   I\models\sm[F;\ {\bf c}] \text{ iff } I\models\sm[F';\ {\bf c}]\ .
\]
\end{thm}

We say that an interpretation is {\em complete} on a predicate $p$ if
it satisfies \hbox{$\forall {\bf x}(p({\bf x})\lor \sneg p({\bf x}))$}.
It is clear that, for any complete interpretation $I$, we have 
$I \models\ \sneg p({\bf t})$ iff $I\models\neg p({\bf t})$.
This fact allows us to use Theorem~\ref{thm:dneg} to replace strong
negation occurring in $H$ with default negation. 

\addtolength{\columnsep}{-20mm}

\medskip\noindent
{\bf Example~\ref{ex:simple} continued}\ \ 
{\it 
Each answer set of the first program in Example~\ref{ex:simple} is
complete. 
In view of Theorem~\ref{thm:dneg}, the first two rules can
be rewritten as\ $p_0\ar \no\ \no\ p_0$\ and\
$\sneg p_0\ar\no\ \no \sneg p_0$,\ which can be further abbreviated as
choice rules $\{p_0\}$ and $\{\sneg p_0\}$. Consequently, the whole program
can be rewritten using choice rules as  \\[-3em]
\begin{multicols}{2}
\[
\small
\ba {c}
  \{p_0\}  \\
  \{\sneg p_0\}  \\ \\
  \{a\} \\
  \{\sneg a\}
\ea
\]

\columnbreak

\[
\small
\ba {c}
  p_1\ar a \\ \\
  \{p_1\} \ar p_0 \\
  \{\sneg p_1\} \ar \sneg p_0 \ . \\ \\
\ea 
\]
\end{multicols} 

Similarly, since $I\models (p_0\mvis\false)$ iff $I\models
\neg(p_0\mvis\true)$, in view of Theorem~\ref{thm:dneg}, the first
rule of the second program in Example~\ref{ex:simple} can
be rewritten as 
\hbox{$p_0\mvis\true \ar \neg\neg (p_0\mvis\true)$}
and further as
$\{p_0\mvis\true\}$. 
This transformation allows us to rewrite the whole program as \\[-3em]

\begin{multicols}{2}
\[
\small
\ba {rcl}
  \{p_0\mvis B\}  \\
  \{a\mvis B\}  
\ea
\]

\[
\small
\ba{rcl}
  p_1\mvis\true &\ar& a\mvis\true \\ 
  \{p_1\mvis B\} &\ar& p_0\mvis B \ ,
\ea 
\]
\end{multicols} 
\noindent
where $B$ ranges over $\{\true,\false\}$. This program represents the
transition system more succinctly than program~\eqref{simple-sneg}.
}
\medskip


\subsection{Representing Strong Negation by Boolean Functions in the
  First-Order Case} \label{ssec:sneg-bool-fo}

Theorem~\ref{thm:sneg-bool-prop} can be extended to the first-order
case as follows. 

Let $f$ be a function constant. A first-order formula is called {\em
  $f$-plain} if each atomic formula  
\begin{itemize}
\item  does not contain $f$, or
\item  is of the form $f({\bf t}) = u$ where ${\bf t}$ is a tuple of
  terms not containing $f$, and $u$ is a term not containing $f$.
\end{itemize}
For example, $f\mvis 1$ is $f$-plain, but each of $p(f)$, $g(f)=1$,
and $1\mvis f$ is not $f$-plain.

For a list $\bC$ of predicate and function constants, we say that a
first-order formula $F$ is $\bC$-plain if $F$ is $f$-plain for each
function constant $f$ in $\bC$.
Roughly speaking, ${\bf c}$-plain formulas do not allow the
functions in ${\bf c}$ to be nested in another predicate or function,
and at most one function in ${\bf c}$ is allowed in each atomic
formula. For example, $f=g$ is not $(f,g)$-plain, and neither is
$f(g)=1\rar g=1$.

Let $F$ be a formula whose signature contains 
both positive and negative predicate constants $p$ and $\sneg p$.
Formula $F^{(p,\sneg\ p)}_{~~b}$ is obtained from $F$ as follows: 
\begin{itemize}
\item  in the signature of $F$, replace $p$ and $\sneg p$ with a new
  intensional function constant $b$ of arity $n$, where $n$ is the
  arity of $p$ (or $\sneg p$), and add two non-intensional object
  constants $\true$ and $\false$;

\item  replace every occurrence of $\sneg p({\bf t})$, where ${\bf t}$ is a
  list of terms, with $b({\bf t})=\false$, and then replace every
  occurrence of $p({\bf t})$ with $b({\bf t})=\true$.
\end{itemize}

By $\i{BC}_b$ (``Boolean Constraint on $b$'') we denote the
conjunction of the following formulas, which asserts that $b$ is a
Boolean function: 
\beq
   \true\ne \false\ , 
\eeq{f1}
\[ 
   \neg\neg\forall {\bf x}(b({\bf x})=\true \lor b({\bf x})=\false)\ ,
\] 
where ${\bf x}$ is a list of distinct object variables.

\begin{thm} \label{thm:sneg-bool-fo}\optional{thm:sneg-bool-fo} 
Let $\bC$ be a set of predicate and function constants, and let $F$ be
a $\bC$-plain formula. Formulas 
\beq
   \forall \bX ((p(\bX) \lrar b(\bX)\mvis\true) 
        \land (\sneg p(\bX) \lrar b(\bX)\mvis\false)),
\eeq{pb}
and $\i{BC}_b$ entail
\[ 
  \sm[F;\ p, \sneg p, \bC] \lrar \sm[F^{(p,\sneg\ p)}_{~~b};\ b,\bC]\ .
\]
\end{thm}

If we drop the requirement that $F$ be $\bC$-plain, the statement does
not hold as in the following example demonstrates. 

\begin{example1}\label{ex:cplain}\optional{ex:cplain}
Take $\bC$ as $(f,g)$ and let $F$ be
$p(f) \land \sneg p(g)$. 
$F^{(p,\sneg\ p)}_{~~b}$ is $b(f)\mvis\true\land b(g)\mvis\false$. 
Consider the interpretation $I$ whose universe is $\{1,2\}$ such that
$I$ contains $p(1), \sneg p(2)$ and with the mappings $b^I(1) = \true,
b^I(2) = \false, f^I = 1, g^I = 2$. $I$ certainly satisfies $\i{BC}_b$ and
(\ref{pb}). $I$ also satisfies $\sm[F;\ p,\sneg p,f,g]$ but does
not satisfy $\sm[F^{(p,\sneg\ p)}_{~~b};\ b,f,g]$: we can let $I$ 
be $\wh{b}^I(1) =
\false, \wh{b}^I(2) = \true, \wh{f}^I = 2, \wh{g}^I = 1$ to
satisfy both $(\wh{b},\wh{f},\wh{g}) < (b,f,g)$ and  
$(F^{(p,\sneg\ p)}_{~~b})^*(\wh{b},\wh{f},\wh{g})$, which is 
$$b(f) = \true \land\wh{b}(\wh{f}) = \true \land b(g) = \false \land
\wh{b}(\wh{g}) = \false.$$
\end{example1}

Note that any interpretation that satisfies both \eqref{pb} and
$\i{BC}_b$ is complete on $p$. Theorem~\ref{thm:sneg-bool-fo} tells us
that, for any interpretation $I$ that is complete on $p$, minimizing
the extents of both $p$ and $\sneg p$ has the same effect as ensuring
that the corresponding Boolean function $b$ have a unique value. 

The following corollary shows that there is a 1--1 correspondence
between the stable models of $F$ and the stable models of 
$F^{(p,\sneg\ p)}_{~~b}$. 
For any interpretation $I$ of the signature of $F$ that is complete on
$p$, by 
$I^{(p,\sneg\ p)}_{~~b}$ we denote the interpretation of the signature of
$F^{(p,\sneg\ p)}_{~~b}$ obtained from $I$ by replacing the relation $p^I$
with function $b^I$ such that 
\[
\ba l
  b^I(\xi_1,\dots,\xi_n)=\true^I \text{\ \  if\ \ }
  p^I(\xi_1,\dots,\xi_n)= \true;  \\
  b^I(\xi_1,\dots,\xi_n)=\false^I \text{\ \ if\ \ } (\sneg
  p)^I(\xi_1,\dots,\xi_n)= \true\  . \\
\ea
\] 
(Notice that we overloaded the symbols $\true$ and $\false$: object
constants on one hand, and truth values on the other hand.) 
Since $I$ is complete on $p$ and coherent, $b^I$ is
well-defined. 
We also require that $I^{(p,\sneg\ p)}_{~~b}$
satisfy~(\ref{f1}). Consequently, $I^{(p,\sneg\ p)}_{~~b}$ satisfies
$\i{BC}_b$. 

\begin{cor}\label{cor:bfelim}\optional{cor:bfelim}
Let $\bC$ be a set of predicate and function constants, and 
let $F$ be a $\bC$-plain sentence.
{\sc (I)} An interpretation $I$ of the signature of $F$ that is complete
on~$p$ is a model of $\sm[F;\ p, \sneg p, {\bf c}]$
iff $I^{(p,\sneg\ p)}_{~~b}$ is a model of 
\hbox{$\sm[F^{(p,\sneg\ p)}_{~~b} ;\ b, {\bf c}]$}.
{\sc (II)} An interpretation $J$ of the signature of $F^{(p,\sneg\ p)}_{~~b}$
is a model of \hbox{$\sm[F^{(p,\sneg\ p)}_{~~b}\land\i{BC}_b ;\ b, {\bf c}]$} 
iff $J = I^{(p,\sneg\ p)}_{~~b}$ for some model $I$ of 
\hbox{$\sm[F;\ p,\sneg p, {\bf c}]$}.
\end{cor}

The other direction, eliminating Boolean intensional functions in
favor of symmetric predicates,  is  similar as we show in the following.

Let $F$ be a $(b,{\bf c})$-plain formula such that every atomic formula
containing $b$ has the form $b({\bf t})=\true$ or $b({\bf t})=\false$,
where ${\bf t}$ is any list of terms (not containing members from
$(b,{\bf c})$). Formula $F^{~b}_{(p,\sneg\ p)}$ is obtained from $F$ as
follows: 
\begin{itemize}
\item  in the signature of $F$, replace $b$ with predicate constants
  $p$ and $\sneg p$, whose arities are the same as that of $b$; 

\item  replace every occurrence of $b({\bf t})=\true$, where ${\bf t}$
  is any list of terms, with $p({\bf t})$, and 
  $b({\bf t})=\false$ with $\sneg p({\bf t})$. 
\end{itemize}

\begin{thm} \label{thm:sneg-boolfunc-pred}\optional{thm:sneg-boolfunc-pred} 
Let $\bC$ be a set of predicate and function constants,
let $b$ be a function constant,  and let $F$ be
a $(b,\bC)$-plain formula such that every atomic formula
containing $b$ has the form $b({\bf t})=\true$ or $b({\bf
  t})=\false$. 
Formulas~(\ref{pb}) and $\i{BC}_b$ entail 
\[ 
  \sm[F;\ b, \bC] \lrar \sm[F^{~b}_{(p,\sneg\ p)};\ p,\sneg p, \bC]\ .
\]
\end{thm}


The following corollary shows that there is a 1--1 correspondence
between the stable models of $F$ and the stable models of 
$F^{~b}_{(p,\sneg\ p)}$. 
For any interpretation $I$ of the signature of $F$ that satisfies
$\i{BC}_b$, by $I^{~b}_{(p,\sneg\ p)}$ we denote the interpretation of the
signature of $F^{~b}_{(p,\sneg\ p)}$ obtained from $I$ by replacing the
function $b^I$ with predicate $p^I$ such that 
\[
\ba l
  p^I(\xi_1,\dots,\xi_n)= \true \text{\ \  iff\ \ } 
     b^I(\xi_1,\dots,\xi_n)=\true^I \ ; \\
  (\sneg p)^I(\xi_1,\dots,\xi_n)= \true\ \text{\ \ iff\ \ }
     b^I(\xi_1,\dots,\xi_n)=\false^I\ .
\ea
\] 

\begin{cor}\label{cor:sneg-boolfunc-pred}\optional{cor:sneg-boolfunc-pred}
Let $\bC$ be a set of predicate and function constants, let $b$ be a
function constant, and 
let $F$ be a $(b,\bC)$-plain sentence such that every atomic formula
containing $b$ has the form $b({\bf t})=\true$ or $b({\bf
  t})=\false$. 
{\sc (I)} An interpretation $I$ of the signature of $F$ is a model
of $\sm[F\land\i{BC}_b;\ b, {\bf c}]$ iff $I^{~b}_{(p,\sneg\ p)}$ is a
model of  \hbox{$\sm[F^{~b}_{(p,\sneg\ p)};\ p,\sneg p, {\bf c}]$}.
{\sc (II)} An interpretation $J$ of the signature of $F^{~b}_{(p,\sneg\ p)}$
is a model of \hbox{$\sm[F^{~b}_{(p,\sneg\ p)};\ p,\sneg p, {\bf c}]$} 
iff $J = I^{~b}_{(p,\sneg\ p)}$ for some model $I$ of 
\hbox{$\sm[F\land\i{BC}_b;\ b, {\bf c}]$}.
\end{cor}

An example of this corollary is shown in the next section.


\section{Representing Non-Boolean Functions Using Strong
  Negation} \label{sec:nonbool-sneg}

In this section, we show how to eliminate non-Boolean intensional
functions in favor of Boolean intensional functions. 
Combined with the method in the previous section, it gives us a
systematic method of representing non-Boolean intensional functions
using strong negation. 



\subsection{Eliminating non-Boolean Functions in Favor of Boolean
  Functions}

Let $F$ be an $f$-plain formula. Formula $F^f_b$ is obtained from $F$
as follows:
\begin{itemize}
\item  in the signature of $F$, replace $f$ with a new boolean
  intensional function $b$ of arity $n+1$ where $n$ is the arity of
  $f$;

\item  replace each subformula $f({\bf t}) = c$ with $b({\bf t},c) =
  \true$.
\end{itemize}

By $\i{UE}_b$, we denote the following formulas that preserve the
functional property:
\[ 
  \forall {\bf x}yz(y\neq z \land b({\bf x},y)=\true \rar  b({\bf
    x},z)=\false), 
\] 
\[ 
   \neg\neg\forall {\bf x} \exists y (b({\bf x},y)=\true),
\] 
where ${\bf x}$ is a $n$-tuple of variables and all variables in 
${\bf x}$, $y$, and $z$ are pairwise distinct.

\BOCC
\begin{remark}
Note that $\i{UE}_b$ is not a constraint as in similar theorems. If we
replace (\ref{UEb1}) with
$$
  \forall {\bf x}yz(y\neq z \land b({\bf x},y)=\true \land b({\bf
    x},z)=\true \rightarrow \bot)
$$
(calling this set $\i{UEC}_b$), we can see that this will not
work. Consider the counterexample $F$ where 
\\$F$ is $f=1$.
\\$F^*$ is $f=1 \land \wh{f} = 1$.
\\$F^f_b$ is $f(1) = \true \land UEC_b$
\\$(F^f_b)^*$ is $f(1) = \true \land \wh{f}(1) = \true \land UEC_b$
\\Clearly the interpretation mapping $f$ to $1$ is a stable model for $F$. However, the analogous interpretation that maps $f(1)$ to $\true$ and $f(x)$ to $\false$ for all $x \neq 1$ is not a stable model of $F^f_b$; we can let $\wh{f}(2)$ be mapped to $\true$ and thus we see the existence of an $\wh{f}$ such that $\wh{f} \neq f$ and $(F^f_b)^*(\wh{f})$ are satisfied.
\end{remark}
\EOCC

\begin{thm}\label{thm:mvfelim}\optional{thm:mvfelim}
For any $f$-plain formula $F$, 
\[
\ba {c}
  \forall {\bf x}y \big(
   (f({\bf x})=y \lrar b({\bf x},y)\mvis\true) 
   \land (f({\bf x})\ne y \lrar b({\bf x},y)\mvis\false)
   \big)
\ea
\]
and $\exists xy(x \neq y)$ entail
\[ 
   \sm[F;\ f, {\bf c}]\ \lrar\ \sm[F^f_b \land\i{UE}_b;\ b, {\bf c}]\ .
\]
\end{thm}

\NB{
\begin{example1}
Note that including 
\[ 
   \forall {\bf x} (b({\bf x}) = \true \lor b({\bf x}) = \false)
\] 
as an assumption is necessary. Consider the formula $F$:
$$f = 1$$
where the universe is $\{1,2\}$. The interpretation mapping $f$ to 1, $b(1)$ to $\true$ and $b(2)$ to 1 satisfies $\sm[F;f]$. However, it does not satisfy $\sm[F^f_b \land\i{UE}_b;b]$ since in particular, it does not satisfy $\i{UE}_b$.
\end{example1}

\begin{example1}
Note that including $\i{UE}_b$ is essential. Consider the formula $F$
$$f = 1$$
where the universe is $\{1,2,3\}$. The unique stable model maps $f$ to $1$. However, if we only consider $F^f_b$, we have
$$b(1) = \true$$
Now, the analogous interpretation mapping $b(1)$ to $\true$ and mapping both $b(2)$ and $b(3)$ to $\false$ is not a stable model since we can map $\wh b(1)$, $\wh b(2)$, and $\wh b(3)$ all to $\true$ to satisfy $\wh b < b \land (F^f_b)^*(\wh b)$.
\end{example1}
}

By $I^f_b$, we denote the interpretation of the signature of $F^f_b$
obtained from $I$ by replacing the mapping $f^I$ with the mapping
$b^I$ such that
\[
\ba{ll}
   b^I(\xi_1,\dots,\xi_n,\xi_{n+1})=\true^I & 
      \text{ if } f^I(\xi_1,\dots,\xi_n) = \xi_{n+1} \\
   b^I(\xi_1,\dots,\xi_n,\xi_{n+1}) = \false^I & 
      \text{ otherwise. }
\ea
\]

\begin{cor}\label{cor:mvfelim}\optional{cor:mvfelim}
Let $F$ be an $f$-plain sentence. 
{\sc (I)} An interpretation $I$ of the signature of $F$ that satisfies
$\exists xy (x\ne y)$ is a model of $\sm[F;\ f,{\bf c}]$ iff $I^f_b$ is a
model of $\sm[F^f_b\land\i{UE}_b;\ b,{\bf c}]$.
{\sc (II)} An interpretation $J$ of the signature of $F^f_b$ that satisfies
$\exists xy (x\ne y)$ is a model of
$\sm[F^f_b\land\i{UE}_b;\ b,{\bf c}]$ iff $J = I^f_b$ for some model
$I$ of  $\sm[F;\ f,{\bf c}]$.
\end{cor}

\noindent
{\bf Example~\ref{ex:bw-func} continued} \ \ 
{\it 
In the program in Example~\ref{ex:bw-func}, we eliminate non-Boolean
function $\i{Loc}$ in favor of Boolean function $\i{On}$ as follows. 
The last two rules are $\i{UE}_{On}$. 
\[
\small
\ba {rcl}
  \bot&\ar&\i{On}(b_1,b,t)\mvis \true\land\i{On}(b_2,b,t)\mvis \true\land b_1 \neq
  b_2 \\
  \i{On}(b,l,t+1)\mvis \true &\ar&\i{Move}(b,l,t) \\
  \bot&\ar&\i{Move}(b,l,t)\land\i{On}(b_1,b,t)\mvis \true \\
  \bot&\ar&\i{Move}(b,b_1,t)\land\i{Move}(b_1,l,t) \\
  \{\i{On}(b,l,0)\mvis \true\}  \\
  \{\i{Move}(b,l,t)\} \\ 
  \{\i{On}(b,l,t+1)\mvis \true\}&\ar&\i{On}(b,l,t)\mvis \true \\
  \i{On}(b,l,t)\mvis \false &\ar& \i{On}(b,l_1,t)\mvis \true \land
  l\ne l_1\\ 
  \bot& \ar &\no\ \exists l (\i{On}(b,l,t)\mvis\true)\ .
\ea 
\]
For this program, it is not difficult to check that the last rule is
redundant. Indeed, since the second to the last rule is the only rule
that has $\i{On}(b,l,t)\mvis\false$ in the head, one can check that
any model that does not satisfy $\exists l (\i{On}(b,l,t)\mvis\true)$
is not stable even if we drop the last rule.


Corollary~\ref{cor:sneg-boolfunc-pred} tells us that this program can
be represented by an answer set program containing strong negation
(with the redundant rule dropped).
\beq
\small
\ba {rcll}
  \bot &\ar&\i{On}(b_1,b,t), \i{On}(b_2,b,t) & (b_1 \neq b_2)\\
  \i{On}(b,l,t+1) &\ar&\i{Move}(b,l,t) \\
  \bot&\ar&\i{Move}(b,l,t), \i{On}(b_1,b,t) \\
  \bot&\ar&\i{Move}(b,b_1,t), \i{Move}(b_1,l,t) \\
  \{\i{On}(b,l,0)\}  \\
  \{\i{Move}(b,l,t)\} \\ 
  \{\i{On}(b,l,t+1)\}&\ar& \i{On}(b,l,t) \\
  \sneg\i{On}(b,l,t) &\ar& \i{On}(b,l_1,t) & (l\ne l_1)\ . \\
\ea 
\eeq{bw-sneg}

Let us compare this program with program~\eqref{bw-sneg0}. 
Similar to the explanation in Example~\ref{ex:simple} (continued), 
the 5th and
the 7th rules of~\eqref{bw-sneg0} can be represented using choice
rules, which are the same as the 5th and the 6th rules
of~\eqref{bw-sneg}. The 6th and the 8th rules of~\eqref{bw-sneg0}
represent the closed world assumption.  We can check that adding these rules to
\eqref{bw-sneg} extends the answer sets of \eqref{bw-sneg0} in a
conservative way with the definition of the negative literals. This
tells us that the answer sets of the two programs are in a 1-1 correspondence.
}\medskip

As the example explains, non-Boolean functions can be represented using
strong negation by composing the two translations, first eliminating non-Boolean
functions in favor of Boolean functions as in
Corollary~\ref{cor:mvfelim} and then eliminating Boolean functions in
favor of predicates as in Corollary~\ref{cor:sneg-boolfunc-pred}. 
In the following we state this composition.


Let $F$ be an $f$-plain formula where $f$ is an intensional function
constant. Formula $F^f_p$ is obtained from $F$ as follows:
\begin{itemize}
\item in the signature of $F$, replace $f$ with two new intensional
  predicates $p$ and $\sneg p$ of arity $n+1$ where $n$ is the arity
  of $f$;

\item replace each subformula $f({\bf t}) = c$ with $p({\bf t},c)$.

\end{itemize}

By $\i{UE}_p$, we denote the following formulas that
preserve the functional property:
\[ 
\ba c
  \forall {\bf x}yz(y\neq z \land p({\bf x},y) \rightarrow\ \sneg
    p({\bf x},z))\ , \\
  \neg \neg \forall {\bf x}\exists y\ p({\bf x},y)\ ,
\ea
\]
where ${\bf x}$ is an $n$-tuple of variables and all variables in
${\bf x},y,z$ are pairwise distinct.

\begin{thm}\label{thm:composition2}\optional{thm:composition2}
For any $(f,\bC)$-plain formula $F$, formulas 
\[  
   \forall {\bf x}y (f({\bf x})= y \lrar p({\bf x},y)), \ \ 
   \forall {\bf x}y (f({\bf x})\ne y \lrar\ \sneg p({\bf x},y)), \ \ 
   \exists xy(x\ne y)
\] 
entail
\[
  \sm[F;\ f,{\bf c}]\lrar\sm[F^f_p\land\i{UE}_{p};\ p, \sneg p, {\bf c}]\ .
\]
\end{thm}

By $I^{~f}_{(p,\sneg\ p)}$, we denote the interpretation of the
signature of $F^{~f}_{(p,\sneg\ p)}$ obtained from $I$ by replacing
the function $f^I$ with the relation $p^I$ that consists of the tuples 
$
  \langle \xi_1,\dots,\xi_n,f^I(\xi_1,\dots,\xi_n)\rangle
$
for all $\xi_1,\dots,\xi_n$ from the universe of $I$. We then also add
the set $(\sneg p)^I$ that consists of the tuples 
$ 
  \langle \xi_1,\dots,\xi_n,\xi_{n+1}\rangle
$
for all $\xi_1,\dots,\xi_n,\xi_{n+1}$ from the universe of $I$ that do
not occur in the set $p^I$.

\begin{cor}\label{cor:composition2cor}\optional{cor:composition2cor}
Let $F$ be an $(f, \bC)$-plain sentence. 
{\sc (I)} An interpretation $I$ of the signature of $F$ that satisfies
$\exists xy(x \neq y)$ is a model of $\sm[F;\ f, {\bf c}]$ iff
$I^{~f}_{(p,\sneg\ p)}$ is a model of 
\hbox{$\sm[F^f_p\land\i{UE}_{p};\ p, \sneg p, {\bf c}]$}. 
{\sc (II)} An interpretation $J$ of the signature of~$F^f_p$ that
satisfies $\exists xy (x\neq y)$ is a model of 
$\sm[F^f_p\land\i{UE}_{p};\ p, \sneg p, {\bf c}]$ 
iff 
$J = I^f_{(p,\sneg\ p)}$ 
for some model $I$ of $\sm[F;\ f, {\bf c}]$.
\end{cor}

Theorem~\ref{thm:composition2} and Corollary~\ref{cor:composition2cor}
are similar to Theorem~8 and Corollary~2
from~\cite{bartholomew12stable}. The main difference is that the
latter statements refer to the constraint called $\i{UEC}_p$ that is
weaker than~$\i{UE}_p$. 
For instance, the elimination method from~\cite{bartholomew12stable}
turns the Blocks World in Example~\ref{ex:bw-func} into almost the same
program as~\eqref{bw-sneg} except that the last rule is turned into the
constraint $\i{UEC}_{On}$: 
\beq
   \ar\i{On}(b,l,t)\land\i{On}(b,l_1,t)\land l\ne l_1 \ .
\eeq{uec-ex}
It is clear that the stable models of $F^f_p\land\i{UE}_p$ are under
the symmetric view, and the stable models of $F^f_p\land\i{UEC}_p$ are
under the asymmetric view. 
To see how replacing $\i{UE}_{On}$ by $\i{UEC}_{On}$ turns the
symmetric view to the asymmetric view, first observe that adding
(\ref{uec-ex}) to program~\eqref{bw-sneg} does not affect the
stable models of the program. Let's call this program $\Pi$. It is easy
to see that $\Pi$ is a conservative extension of the program that is
obtained from $\Pi$ by deleting the rule with $\sneg\i{On}(b,l,t)$ in
the head.


\section{Relating to Lifschitz's Two-Valued Logic Programs}
\label{sec:tvlp}

~\citeauthor{lifschitz12two-valued}~[\citeyear{lifschitz12two-valued}]
presented a high level definition of a logic program that does not
contain explicit default negation, but can handle nonmonotonic
reasoning in a similar style as in Reiter's default logic. In this
section we show how his formalism can be viewed as a special case of
multi-valued propositional formulas under the stable model semantics
in which every function is Boolean. 



\subsection{Review: Two-Valued Logic Programs} 

Let $\sigma$ be a signature in propositional logic. 
A {\em two-valued rule} is an expression of the form 
\beq
  L_0 \ar L_1, \dots, L_n : F
\eeq{liftvrule}
where $L_0,\dots,L_n$ are propositional literals formed from $\sigma$
and $F$ is a propositional formula of signature $\sigma$. 

A {\em two-valued program} $\Pi$ is a set of two-valued rules. 
An interpretation $I$ is a function from $\sigma$ to
$\{\true,\false\}$. 
The {\em reduct} of a program $\Pi$ relative to an interpretation $I$,
denoted $\Pi^I$, is the set of rules 
$ 
  L_0 \ar L_1,\dots,L_n
$ 
corresponding to the rules~(\ref{liftvrule}) of $\Pi$ for which 
$I\models F$. Interpretation $I$ is a stable model of $\Pi$ if it is a
minimal model of $\Pi^I$. 

\begin{example1}\label{ex:tv}
\beq
\ba l
  a ~\ar~ :\ a, \qquad  \neg a  \ar  :\ \neg a, \qquad 
  b  \ar  a\ :\ \top
\ea
\eeq{tv-example}
The reduct of this program relative to $\{a,b\}$ consists of rules $a$
and $b\ar a$. Interpretation $\{a,b\}$ is the minimal model of the
reduct, so that it is a stable model of the program.
\end{example1} 

As described in \cite{lifschitz12two-valued}, if $F$ in every
rule~(\ref{liftvrule}) has the form of conjunctions of literals, then
the two-valued logic program can be turned into a traditional answer
set program containing strong negation when we consider complete
answer sets only. For instance, program~(\ref{tv-example}) can be
turned into
\[
\ba l 
   a ~\ar~ \no\ \sneg a,  \qquad
   \sneg a ~\ar~ \no\ a, \qquad
   b ~\ar~ a\ .
\ea
\]
This program has two answer sets, $\{a,b\}$ and ${\sneg a}$, and only
the complete answer set $\{a,b\}$ corresponds to the stable model
found in Example~\ref{ex:tv}.


\subsection{Translation into SM with Boolean Functions}

Given a two-valued logic program $\Pi$ of a signature $\sigma$, 
we identify $\sigma$ with the multi-valued propositional signature 
whose constants are from $\sigma$ and the domain of every constant is
Boolean values $\{\true,\false\}$. 
For any propositional formula $G$, $\i{Tr}(G)$ is obtained from~$G$ by
replacing every negative literal $\sneg A$ with $A\mvis\false$ and
every positive literal $A$ with $A\mvis\true$. 
By $\i{tv2sm}(\Pi)$ we denote the multi-valued propositional formula
which is defined as the conjunction of 
\[
  \neg\neg\i{Tr}(F)\land\i{Tr}(L_1)\land\dots\land\i{Tr}(L_n)\rar\i{Tr}(L_0) 
\]
for each rule~(\ref{liftvrule}) in~$\Pi$.



For any interpretation $I$ of $\sigma$, we obtain the multi-valued
interpretation $I'$ from $I$ as follows. For each atom
$A$ in $\sigma$,
\begin{displaymath}
   I'(A) = \left\{
     \begin{array}{ll}
       \true \hspace{2em} & \text{if $I\models A$} \\
       \false & \text{if $I\models \neg A$} 
     \end{array}
   \right.
\end{displaymath} 

\begin{thm}\label{thm:liftvtrans} \optional{thm:liftvtrans}
For any two-valued logic program $\Pi$, an
interpretation $I$ is a stable model of $\Pi$ in the sense of 
\cite{lifschitz12two-valued} 
iff $I'$ is a stable
model of $\i{tv2sm}(\Pi)$ in the sense of \cite{bartholomew12stable}.
\end{thm}


\medskip\noindent
{\bf Example~\ref{ex:tv} continued}\ \
{\it 
For the program $\Pi$ in Example~\ref{ex:tv}, $\i{tv2sm}(\Pi)$ is the
following multi-valued propositional formula: 
\[
\small
\ba l
  ~~~\big( \neg\neg (a\mvis \true) \rar a\mvis \true \big) 
  \land\ \big( \neg\neg (a\mvis \false) \rar a\mvis \false \big) 
  \land\ \big( a\mvis \true \rar b\mvis \true \big).
\ea
\]
According to~\cite{bartholomew12stable}, this too has only one stable
model in which $a$ and $b$ are both mapped to $\true$, corresponding
to the only stable model of $\Pi$ according to Lifschitz. 
}



Consider extending the rules \eqref{liftvrule} to contain
variables. It is not difficult to see that the translation
$\i{tv2sm}(\Pi)$ can be straightforwardly extended to non-ground
programs. This accounts for providing the semantics of the first-order
extension of two-valued logic programs.

\section{Strong Negation and the Cabalar
  Semantics} \label{sec:sneg-cabalar}

There are other stable model semantics of intensional
functions. Theorem~5 from~\cite{bartholomew13onthestable} states that
the semantics by
\citeauthor{bartholomew13onthestable}~[\citeyear{bartholomew13onthestable}]
coincides with the semantics by 
\citeauthor{cabalar11functional}~[\citeyear{cabalar11functional}]
on $\bC$-plain formulas. Thus several theorems in this note stated for
the Bartholomew-Lee semantics hold also under the Cabalar semantics.

A further result holds with the Cabalar semantics since it allows
functions to be partial. This provides extensions of
Theorem~\ref{thm:sneg-bool-fo} and Corollary~\ref{cor:bfelim}, which do
not require the interpretations to be complete. Below we state this
result. Due to lack of space, we refer the reader to
\cite{bartholomew13onthestable} for the definition of $\cbl$,
which is the second-order expression used to define the Cabalar
semantics.


Similar to $\i{BC}_b$ in Section~\ref{ssec:sneg-bool-fo}, by
$\i{BC}'_b$ we denote the
conjunction of the following formulas:
\beq
   \true\ne \false, 
\eeq{fc1}
\[ 
   \neg\neg\forall {\bf x}(b({\bf x})=\true \lor b({\bf x})=\false\lor
   b({\bf x})\ne b({\bf x})), 
\] 
where ${\bf x}$ is a list of distinct object variables.\footnote{Under
  partial interpretations, $b({\bf t})\ne b({\bf t})$ is true if
  $b({\bf t})$ is undefined. See~\cite{cabalar11functional,bartholomew13onthestable} for
  more details.} 


\begin{thm} \label{thm:sneg-bool-fo-c}\optional{thm:sneg-bool-fo} 
Let $\bC$ be a set of predicate constants, and let $F$ be a formula. 
Formulas 
\[ 
   \forall \bX ((p(\bX) \lrar b(\bX)\mvis\true) 
        \land (\sneg p(\bX) \lrar b(\bX)\mvis\false)
        \land (\neg p(\bX)\land \neg\!\sneg p(\bX)\lrar b(\bX)\ne b(\bX)),
\] 
and $\i{BC}'_b$ entail \footnote{{\rm The entailment is under partial
  interpretations and satisfaction.}}
\[ 
  \sm[F;\ p, \sneg p, \bC] \lrar \cbl[F^{(p,\sneg\ p)}_{~~b};\ b,\bC]\ .
\]
\end{thm}

The following corollary shows that there is a 1--1 correspondence
between the stable models of $F$ and the stable models of 
$F^{(p,\sneg\ p)}_{~~b}$.\footnote{Recall the notation defined in
  Section~\ref{ssec:sneg-bool-fo}.} 
For any interpretation $I$ of the signature of $F$, by 
$I^{(p,\sneg\ p)}_{~~b}$ we denote the interpretation of the signature of
$F^{(p,\sneg\ p)}_{~~b}$ obtained from $I$ by replacing the relation $p^I$
with function $b^I$ such that 
\[
\ba l
  b^I(\bfxi)=\true^I \text{\ \  if\ \ }
  p^I(\bfxi)= \true\ ;  \\
  b^I(\bfxi)=\false^I \text{\ \ if\ \ } (\sneg
  p)^I(\bfxi)= \true\ ;    \\
  b^I(\bfxi)=u \text{\ \ if\ \ }  
    p^I(\bfxi)= 
    (\sneg p)^I(\bfxi)= \false\  . 
\ea
\] 
Since $I$ is coherent, $b^I$ is
well-defined. 
We also require that $I^{(p,\sneg\ p)}_{~~b}$
satisfy~(\ref{fc1}). Consequently, $I^{(p,\sneg\ p)}_{~~b}$ satisfies
$\i{BC}'_b$. 

\begin{cor}\label{cor:bfelim-c}\optional{cor:bfelim-c}
Let $F$ be a sentence, and let $\bC$ be a set of predicate constants.
{\sc (I)} An interpretation $I$ of the signature of $F$ is a model of
$\sm[F;\ p, \sneg p, {\bf c}]$
iff $I^{(p,\sneg\ p)}_{~~b}$ is a model of 
\hbox{$\cbl[F^{(p,\sneg\ p)}_{~~b} ;\ b, {\bf c}]$}.
{\sc (II)} An interpretation $J$ of the signature of $F^{(p,\sneg\ p)}_{~~b}$
is a model of \hbox{$\cbl[F^{(p,\sneg\ p)}_{~~b}\land\i{BC}'_b ;\ b, {\bf c}]$} 
iff $J = I^{(p,\sneg\ p)}_{~~b}$ for some model $I$ of 
\hbox{$\sm[F; p,\sneg p, {\bf c}]$}.
\end{cor}


\section{Conclusion} \label{sec:conclusion}

In this note, we showed that, under complete interpretations,
symmetric predicates using strong negation can be alternatively
expressed in terms of Boolean intensional functions in the language of
\cite{bartholomew12stable}. They can also be expressed in
terms of Boolean intensional functions in the language of
\cite{cabalar11functional}, but without requiring the complete
interpretation assumption, at the price of relying on 
the notion of partial interpretations. 
%

System {\sc cplus2asp}~\cite{casolary11representing,babb13cplus2asp}
turns action language ${\cal C}$+ into answer
set programs containing asymmetric predicates. The translation in this
paper that eliminates intensional functions in favor of symmetric
predicates provides an alternative method of computing ${\cal C}$+
using ASP solvers. 



\medskip
\noindent
{\bf Acknowledgements: } 
We are grateful to Vladimir Lifschitz for bringing attention to this
subject, to Gregory Gelfond for useful discussions related to this
paper, and to anonymous referees for useful comments. 
This work was partially supported
by the National Science Foundation under Grant IIS-0916116 and by the
South Korea IT R\&D program MKE/KIAT 2010-TD-300404-001.

\bibliographystyle{named}


\end{document}